SciDoc Publishers
Infer, Interpret & Inspire Science

**International Journal of Mechatronics and Automotive Research (IJMAR)**

# A Framework for Controlling Multiple Industrial Robots using Mobile Applications

Research Article

Daniela Alvarado[1], Dr. Seemal Asif[2*]

[1] School of Aerospace, Transport and Manufacturing, Cranfield University, England.
[2] Centre for Structures, Assembly and Intelligent Automation, School of Aerospace, Transport and Manufacturing, Cranfield University, England.

**Abstract**

**Purpose:** Over the last few decades, the development of the hardware and software has enabled the application of advanced systems. In the robotics field, the UI design is an intriguing area to be explored due to the creation of devices with a wide range of functionalities in a reduced size. Moreover, the idea of using the same UI to control several systems arouses a great interest considering that this involves less learning effort and time for the users. Therefore, this paper will present a mobile application to control two industrial robots with four modes of operation.

**Design/methodology/approach:** The smartphone was selected to be the interface due to its wide range of capabilities and the MIT Inventor App was used to create the application, whose environment is supported by Android smartphones. For the validation, ROS was used since it is a fundamental framework utilised in industrial robotics and the Arduino Uno was used to establish the data transmission between the smartphone and the board NVIDIA Jetson TX2. In MIT Inventor App, the graphical interface was created to visualize the options available in the app whereas two scripts in python were programmed to perform the simulations in ROS and carry out the tests.

**Findings:** The results indicated that the use of the sliders to control the robots is more favourable than the Orientation Sensor due to the sensibility of the sensor and human limitations to hold the smartphone perfectly still. Another important finding was the limitations of the autonomous mode, in which the robot grabs an object. In this case, the configuration of the Kinect camera and the controllers has a significant impact on the success of the simulation. Finally, it was observed that the delay was appropriate despite the use of the Arduino UNO to transfer the data between the Smartphone and the Nvidia Jetson TX2.

**Originality/value:** The following points show the contributions of this paper to the robotic field.
• Developed a robust application combining four different modes of functionality.
• Create a program with an intuitive interface and simple use that allows controlling different robotics arms with just one UI.
• Evaluate the applicability of the simulated industrial robots in ROS since the projects must be tested in a framework before implementing it in real robots.

## Introduction

In the history of the development of technological advances, robots have been thought to be the key factor in the improvement of people's life quality. In general, it is used to perform repetitive tasks and to manipulate harmful substances in industries. However, over the last decades, the advances in the hardware and software have allowed us to use them in diverse environments without the necessity of being in a cage since it is considered that robots are safe enough to interact directly with humans. Moreover, the development of vision systems, components such as microphones and sensors; and the advances of Machine Learning have renewed the interest to incorporate vision, speech, and voice recognition capabilities in robots. As well as, there has been an increasing interest in the development of autonomous robots and the design of versatile interfaces whose objective is to increase workers' trust in them and encourage people to use these systems in their daily life. As a result, many companies are currently developing new technology, and prototypes of autonomous vehicles

**Citation:** Daniela Alvarado, Dr. Seemal Asif. A Framework for Controlling Multiple Industrial Robots using Mobile Applications. *Int J Mechatron Autom Res.* 2021;3(1):13-18.

can be seen travelling on the streets of many cities. An example is the Starship robot that delivers items in Milton Keynes, whose number of delivery services has increased due to the COVID-19. This is a clear example of how robots can improve and make easier people's lives.

Nowadays, researchers can experiment and develop their systems since there are several resources available for users. One interesting area has been the introduction of the smartphone as a UI, which has been used in many technical articles. Although there are many projects oriented to control kits of robotic arms, the approach of this framework is to control multiple industrial robots with the same UI. Being the prototype of an innovative, intuitive, and robust application that includes low-cost and serviceable options such as its compatibility with a widely Android platform which is an advantage over the current expensive control mechanisms. Another economic benefit of using ARI is the training cost of the workers since it could be reduced by using a single instead of several UI in controlling of the multiple industrial robots. Moreover, ARI is a key contributor to Industry 4.0 and IoTas it will improve the interconnectivity and communication between the different systems. As well as, it could make easier the reporting, monitoring, and the collection of the data improving its integration into the central manufacturing ERP system.

It is known that one issue in the area of technology is that almost each device OS has a different programming interface. According to Delden S. and Whigham A., the use of a smartphone app to control an articulated robot can reduce the user learning effort and non-expert user can interact with the system [17].

Shelvam M. suggested that the implementation of the smartphone reduces the programming time, makes a user-friendly interface, and encourages the progress of other electronic devices. In his experiment, the HC-05 Bluetooth module and the App Inventor were used to control a surveillance robot [15].

Similar to the previous project, Shoeb M. and Borole P. suggested that the implementation of a smartphone in a surveillance robot can reduce the risk of injury for humans [16]. In this case, the device captures the audio and video and it is transferred via Bluetooth to the tablet or laptop. As a result, the user can get live information and send the respective commands to control the robot.

It was pointed out above, several studies have been performed to assess the efficacy of a smartphone to capture video in real-time. In the following two studies, the App Inventor was implemented to control the robot and the user could observe the environment using the video feed from the smartphone located in the robot. In the first project, the Mobizen environment was implemented to emulate an Android device in the PC [1]. In the second one, the environment Eclipse was chosen to compile the project and the S5PV210 embedded board was implemented to control the mechanism of the hexapod robot [18].

Furthermore, previous research suggested that the incorporation of a smartphone reduces the workspace needed and it is a low-cost solution [8, 14].

As was mentioned previously, a smartphone has incorporated many components and one of them is a microphone, which allows performing experiments with voice recognition. Evidence of this statement, a project with a robotic arm was presented, in which the motion is controlled with this feature [11]. However, other authors suggested that this method requires a special microphone to capture a high-quality signal, which limits its implementation and accessibility [8].

In the following project, the joints were controlled by the accelerometer and gyroscope of the smartphone and the data transmission was via WIFI using the IEEE802.11n standard. The two main functions of the program are the calculation of the IK and the signal filtering to reduce the noise from the sensor. The results showed that it has limitations due to the response time and the high sensitivity of the sensors [12, 13].

So far, this paper has focused on the control of the manipulator using a single method. The following paragraphs will show two projects, in which a combination of two methods was performed. In the first one, the Lego Mindstorms and the App Inventor were used. According to the application, the system could be controlled by using the touch screen and voice recognition. Firstly, the application was developed in Eclipse and the data transmission was via Bluetooth. Secondly, the voice recognition was processed via the Internet, which was executed on Google's servers [10].

In the second project, the robotic arm and grippers were controlled by voice, while the motion of the body was performed by the tilt gesture of the smartphone [9]. Regarding the voice recognition, an on-board module was used to convert the speech input into logical signals, in which the commands were associated with determinate stored keywords. The results showed that the application was quite precise, although it was suggested to use a more robust signal processing to reduce the background noise.

Before proceeding to introduce the next section, it is important to present the utilization of the board Jetson TX2 in previous projects.

Previous studies have explored the implementation of ROS in the board Nvidia Jetson TX2, in which most of the projects it was used to develop mobile systems. In a recent experiment, this board and ROS were used to accomplish an autonomous navigation robot, in which the TensorRT and OpenCV were implemented to achieve the objective [6].

Moreover, it is known that the robots are very expensive systems; hence it was suggested to recreate a variation of the Turtlebot2 design in the real world. To do that, the parts of the robot were 3D printed and the hardware components were assembled by the researchers. However, the Next Unit of Computing (NUC) controller was used instead of the Jetson TX2 due to its volume. Other components implemented in this system were an Arduino UNO, DC motor, and drivers; whereas ROS with the SLAM and Rviz framework was used to achieve the goal [7].

As has been observed, many projects had been carried out, in which the main objective was to develop low-cost systems to control consumer robotic arms. However, there is still a challenge in the industry because each system is controlled by its own interface, which usually is expensive and complex. Hence, the innovation of this framework is to control multiple industrial robots using an accessible and inexpensive UI.

## Framework Setup

The framework has been developed and the virtual simulations performed with real robot models, which support the idea of using this app in industrial applications. Figure 1 shows the architecture diagram of the system, in which the Bluetooth connection must be established; as well as, selecting the robot and the mode of operation. The numbers "5000" and "5001" are used as tags to indicate the robot selected. Based on the robot chosen, its number and the values set in the operation interface are sent from the smartphone to the Arduino UNO via Bluetooth in 2 bytes, whose value is set into a variable of 16 bits. Moreover, the tags are converted to strings, reducing the risk of misappropriation of the numbers in the script code. After that, the parameter is transferred to the Nvidia Jetson TX2 through the rosserial connection, where the values are set into the respective nodes in Rviz and Gazebo. Finally, the robot's motion is visualized setting the parameters into the controllers.

The developed application is named as "ARI", which means Android Robotic Interface, and has four modes of operation.

• Mode 1: Control each joint using sliders. The user controls the manipulator using sliders, providing the values to rotate the joints and move the gripper fingers.
• Mode 2: Send the target position. Seven parameters must be set including the coordinates and orientation of the robot target position.
• Mode 3: Tilt gesture. An Orientation Sensor is used to provide the roll angle of the Gyroscope, whose values are used to rotate the joints. Two sliders are used to move the fingers of the gripper.
• Mode 4: Autonomous mode - Grab an object. Two buttons are available to start and stop the motion of the manipulator. Moreover, the information about the state of the simulation is shown on the screen of the smartphone.

To begin with, the framework in Ros was performed using the world model developed by the finalist of the ARIAC competition of 2017, which has the suitable features of an industrial environment as it can be seen in Figure 2 [2]. After that, the models of the Panda Emika and UR5 robots were cloned from the repository GitHub and spawned into the framework [3, 5]. The UR5 model provided in the universal_robot package does not have the gripper, so the gripper model of the Panda robot was attached to the UR5 due to its operational simplicity.

Then two MoveIt packages were created to configure the controllers and parameters such as the planning groups and poses of each robot. The second MoveIt package was created for the configuration of the controllers in the fourth, whose frame was different from the other three. Therefore, it was decided to set effort controllers to apply enough force through the gripper to grab the object, whereas position controllers were set in the first MoveIt package to control the joints in the other three modes of operation.

In the MIT Inventor App, the graphic interfaces of the four modes of operation were designed. Whereas the Arduino Software was used as an interface between the smartphone and the NVIDIA Jetson TX2, to convert the numbers used as a tag in MIT Inventor App to a string, reducing the risk of misappropriation of numbers in the script code. Moreover, two scripts were coded following the Object-Oriented Programming (OOP) template. In the first script, the functions of the three first modes of operation were programmed. The same function is used for the first and third modes since their structures are similar. Therefore, the string tag received from the Arduino UNO is associated with their respective joint, and the parameter received to rotate the joint or move the fingers of the gripper is set to visualize the simulation in Rviz and Gazebo. Regarding the second mode, the seven parameters (x, y, z, Xor, Yor, Zor, W, and Confirm) are set to plan the new target.

On the other hand, a package created by Mickel Ferguson was cloned from Github to perform the fourth mode of operation (4). The script provided in this package was adapted for this project. A Kinect camera was implemented to detect the objects and provide the position information to the robot to calculate the IK. This information is updated every 5 seconds, and any objects placed in 1.85m are detected as long as it is on a surface elevated 0.5m from the ground. Moreover, a cube is created to simulate the body of the table as illustrated in Figure 3, avoiding thus the collision of the robot with the table.

**Figure 1. Architecture diagram of ARI**

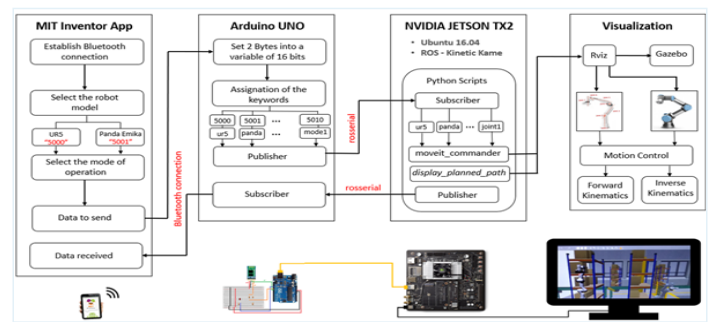

**Figure 2. Gazebo - Simulation**

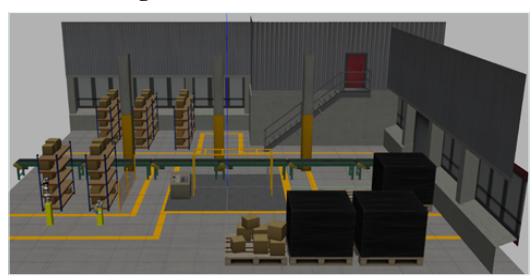

**Modes of Operation**

Figure 4 shows the starting screen that is shown when the user opens ARI. Once the Bluetooth connection has been established, the user must select the robot and the mode of operation.

**Mode 1: Control each joint using sliders:** Eight sliders were used to control the six joints, whose range was from -180° to 180°, and two for the gripper, whose maximum and minimum values were 4cm and 0cm. The setup is shown in Figure 5. This first mode is based on the FK since the user controls the joints by sending the values of the angles to be rotated. Because it is not possible to send negative numbers, a loop was defined in which if the number is negative, the value is multiplied by (-1) and the number 1000 is added.

**Mode 2: Send the target position:** In this case, the parameter set for the user is multiplied by 100 since the units used are in meters and it is not possible to send float numbers. The setup of mode 2 is represented in Figure 6. According to this mode, the user must select firstly the coordinate or rotation button, which is associated with a determined tag. Then, the number written in the text box is sent by selecting the right button, and finally, to confirm the target position, the "Confirm" button must be pushed.

**Mode 3: Tilt gesture:** The Orientation Sensor available in MIT Inventor App was used in this mode; which provides the Roll, Yaw, and Pitch positions obtained from the gyroscope. The setup of mode 3 is shown in Figure 7. In this case, the user must push the button of the joint to be controlled, making the Roll angle visible on the text box, and then the value is sent when the user pushes the sent button. Due to the instability of the sensor and the small range of values allowed in the motion of the grippers; it was decided to control the fingers using two sliders like in mode 1.

**Mode 4: Autonomous mode – Grab an object:** There are two buttons available on the screen as it can be seen in Figure 8, which are used to start and stop the simulation. Moreover, two switches were included to show the activation of these buttons. The main aspect of this mode is that the information provided during the simulation is shown on the screen using four switches. The function "Timer" available in MIT Inventor App is applied to provide this information regularly using the internal clock of the smartphone.

## Results

The functionality, sensibility, and latency of the system were evaluated to prove its applicability in industrial applications.

**Functionality**

The functionality of ARI was measured according to the accuracy and precision of the different modes of operation. The method applied to determine the accuracy was to calculate the MAE of the data, whereas the standard deviation was applied to measure the dispersion of these values. The result showed that the first and third modes of operation are the most accurate and precise. Since their MAE and standard deviation have the smallest value.

**Figure 3. Perception of the Kinect Camera**

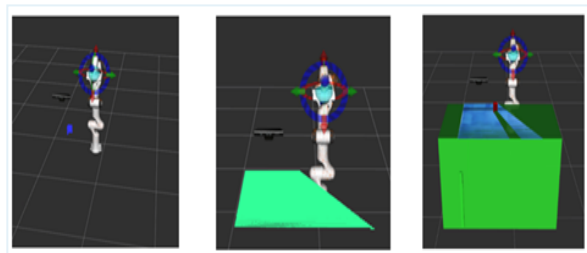

**Figure 4. Starting screen - To establish the Bluetooth connection and select the robot and mode of operation**

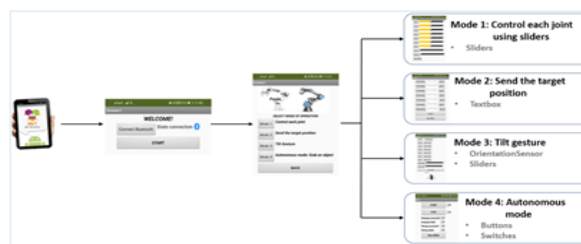

**Figure 5. Operation of Mode 1**

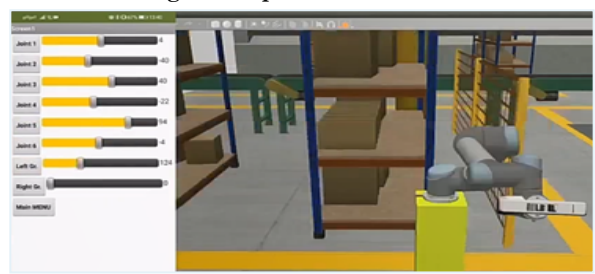

Whereas the last mode of operation has the least functionality. Table 1 shows the data collected around all the modes to test the functionality of the framework and Figure 9 illustrated the results for all four modes.

**Sensibility**

Regarding the sensibility, the joints and gripper have a sensibility of 1º and 1mm, respectively. Being these values suitable to be used in industrial applications. Another aspect to take into account is the human limitations to hold an object perfectly still, which makes it difficult to send the desired value using the OrientationSensor. Whereas the sensibility of the fourth mode of operation is closely related to the Kinect Camera and the solution provided by the IK solver. If the robot spent too much time trying to grasp the object and some part of it is in the area of perception of the camera, that part of the robot will be detected as an object. On the other hand, some solutions provided for the IK are complex and as a result, the robot has an unusual and strange motion. This causes the robot to drop the object during the simulation. As a result, several attempts are required to achieve a successful simulation. In this case, only 55% of the time the robot placed the object on the table.

**Delay**

The last test was to measure the delay. So, the time spent when a publisher in ROS sends a value to the Arduino UNO through ros serial, which is a protocol for allowing the communication between ROS and a character device; and it sends a response to

ROS was obtained. It was 32.8ms, which was divided into two. As a result, the time spent in the data transmission was 16.4ms, being this value appropriate for a serial connection between two devices. Moreover, a function was used to measure the time from the first byte received from the Bluetooth module to when the value was sent to ROS, which was 2ms, being this value suitable for a code in Arduino. Finally, it was evaluated the real-time factor in Gazebo, which means the synchronization between the real-time and the simulation. The expected value should be between 0.9 and 1, but it was 0.7, reducing the speed of the simulation by 30%. This reduction is due to various factors such as the number of elements spawned in the Gazebo, the model of the world used, and the physics and features of the elements implemented in the simulation, among others.

## Conclusion

A framework was developed to control the UR5 and Panda robot using a mobile app with four modes of operation, in which different sensors and tools were implemented such as sliders and the OrientationSensor, among others. Once ARI was created, the functionality, sensibility, and delay were evaluated to validate its applicability in the industry. The results related to the functionality shown that the first and third modes of operation are the most accurate; being these the best options to be used in those tasks where high precision is required. Regarding the second mode of operation, in some samples the absolute error was 1cm. Being this value not suitable to be applied in industrial applications. However, this error could be reduced modifying the sensibility of the system. In this case, the number provided by the user is multiplied by 100; but if this was multiplied by 1000, the sensibility could be of three decimals, reducing thus the absolute error of the system. The results of the test of the last mode of operation indicated that it is required many attempts to achieve a successful simulation since 45% of the time the robot failed in the delivery process. Moreover, the results of the third test support the idea that the implementation of the Arduino UNO as an interface increases the delay slightly. Nevertheless, the delay in the data transmission is low, being this value appropriate to control manipulators in industrial applications.

**Figure 6. Operation of Mode 2**

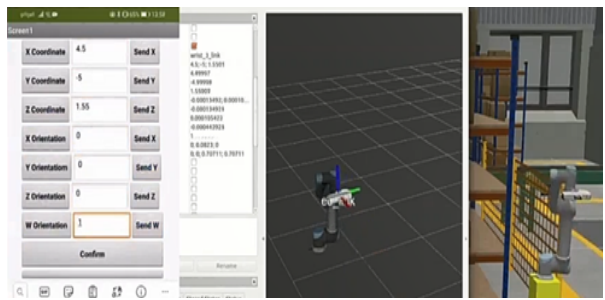

**Figure 7. Operation of Mode 3**

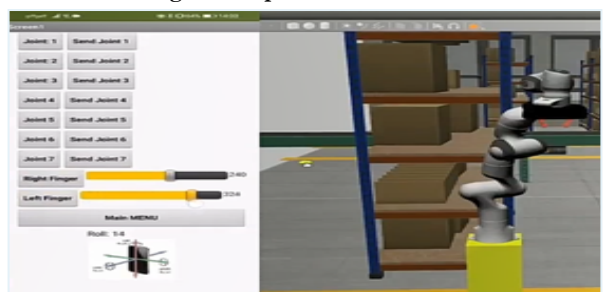

**Figure 8. Operation of Mode 4**

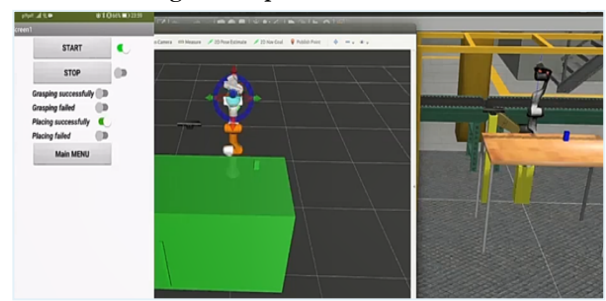

**Table 1. Functionality**

| | N° Samples | Accuracy MAE | 40 Samples | Precision S. Deviation | 40 Samples |
|---|---|---|---|---|---|
| Mode 1 | 44 | 0.0000556 | 0.0000539 | 0.0000849 | 0.0000853 |
| Mode 2 | 15 3 = 45 | 0.00109 | 0.001175 | 0.00286 | 0.002815 |
| Mode 3 | 42 | 0.0000528 | 0.0000526 | 0.0000679 | 0.0000683 |
| Mode 4 | 20 2 = 40 | 0.19325 | 0.19325 | 0.308355 | 0.308355 |

**Figure 9. MAE and Standard deviation**

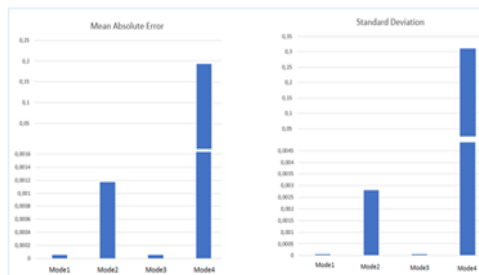

On the other hand, ARI is safe enough since the first three modes of operation work only when the user is controlling the system. Whereas the autonomous mode has a "Stop" button, to move back the robot to the starting position if there is some issue during the simulation. However, it is suggested to explore and analyze the security patches and policies to make ARI follow the security legislation and regulations which are implemented in the industry. Then, ARI could be suitable for a wide number of functionalities and environments. For example, for industrial applications where the robots must do repeated tasks. Universities and research centers to develop the next generation of robots. As well as, it could be used for people with limited movement capabilities, improving their life quality.